\documentclass{article} % For LaTeX2e
\usepackage{retro_style}
\usepackage{enumitem}

\usepackage{microtype}
\usepackage{hyperref}
\usepackage{url}
\usepackage{booktabs}
\usepackage{listings}
\usepackage{multirow}
\usepackage{graphicx}
\usepackage{subcaption}
\usepackage{multirow} % Multi rows in tables
\usepackage{makecell} % Multi cells in tables
\usepackage{arydshln} % Dashed lines in tables
\usepackage{bold-extra} % Bold face inside texttt environment

\usepackage{pifont}% http://ctan.org/pkg/pifont

\usepackage{fontspec}
\retroheader{WorkBench Revisited}

\title{WorkBench Revisited: Workplace Agents Two Years On}

\author{Olly Styles \\
Independent researcher \\
\texttt{ollystyles@gmail.com}
\And
Sam Miller \\
Independent researcher \\
}

\colmfinalcopy % Show the author block (this is not a blind submission).
\begin{document}

\maketitle

\begin{abstract}
The best agent on WorkBench in March 2024, GPT-4, completed just 43\% of tasks. We revisit the benchmark in June 2026 and find that the best agent to date, Claude Fable 5, now completes 98\%. Beyond this considerable progress in frontier agent performance, three things stand out. First, unintended harmful actions, such as emailing the wrong person, fell from 26\% of tasks for GPT-4 to 1.9\% for Claude Fable 5; capability and safety go together on WorkBench rather than trade off, so the models that finish the most tasks also do the least unintended damage. Second, the rise of open-weight models has drastically lowered costs for a performance level that was only accessible to proprietary models, while frontier costs have stayed stable. Third, while several classes of error have been eliminated, frontier models still make some basic mistakes that occasionally result in irreversible harm. We release an updated version of the benchmark with data and code quality improvements, new model scores, and analysis of agent progress on WorkBench since 2024.
\end{abstract}

\begin{center}
\includegraphics[width=\linewidth]{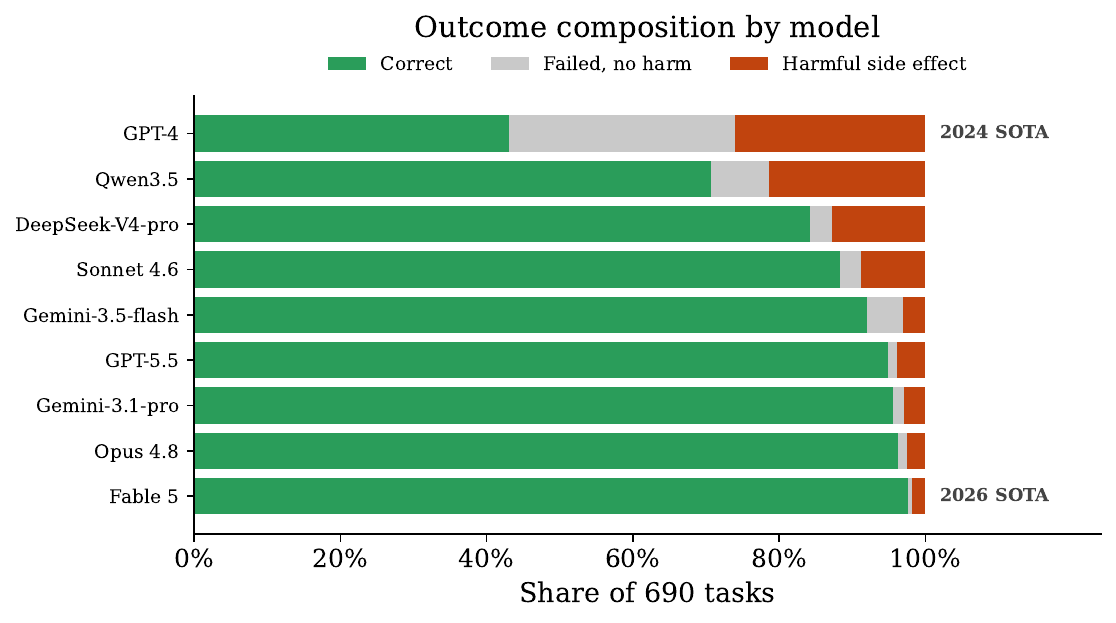}
\captionsetup{hypcap=false}
\captionof{figure}{\textbf{Outcome composition by model.} Each model's 690 WorkBench
tasks split into correct, failed-but-harmless, and harmful side effects, ordered by
task completion (best at bottom). GPT-4 is the original 2024 result (43\% completion,
26\% side effects, scored with a ReAct loop on the pre-revision benchmark); the other
eight are 2026 runs from Table~\ref{tab:model-results}.}
\label{fig:outcome-composition}
\end{center}

\section{Introduction}
\label{sec:intro}
Large language models are increasingly deployed as agents that take actions on a
user's behalf: updating customer records, managing email, scheduling meetings. Most
agent benchmarks measure something adjacent to this, such as web navigation
\citep{zhou2024webarena}, general assistance \citep{mialon2023gaia}, or broad tool
use \citep{liu2023agentbench}. WorkBench \citep{styles2024workbench} was built to
measure the workplace use case instead. It provides a sandbox environment of five databases (a
calendar of 300 events, an inbox of 500 emails, 500 website-analytics records, a
customer relationship manager with 200 customers, and a project-management board of
300 tasks), 26 tools to read from and write to, and 690 tasks generated from templates
(10 tasks per template) that span single-domain and multi-domain work. A task is graded by comparing the sandbox's
final state against the ground truth, so the agent can take any path it likes,
recover from its own mistakes, and there's no second model sitting in judgment. This sets WorkBench apart from action-matching benchmarks and from those that
lean on an LLM evaluator, such as ToolLLM \citep{qin2023toolllm}.

At release in 2024 the benchmark was far from solved. The strongest agent, a ReAct
\citep{yao2022react} loop around GPT-4, completed 43\% of tasks; the weakest open
model evaluated, Llama2-70B, managed 3\%. Two years later, the set of available
models is almost entirely new. This paper examines what has changed, and why.

We do three things:
\begin{itemize}[leftmargin=*]
\itemsep0.2em
  \item We re-run WorkBench on 24 models released between 2023 and 2026, spanning
  eight vendors and both proprietary and open-weight models, under a single modern
  agent harness that uses native tool-calling rather than text-parsed ReAct
  (Section~\ref{sec:setup}).
  \item We report task completion alongside a new axis the original paper did not report: dollar cost of completing each task (Section~\ref{sec:results}).
  \item We correct a set of scoring, ground-truth, and prompt issues in the
  original benchmark, make engineering improvements to its tools, and quantify which
  tasks they affect, so that results on the 2026 benchmark are comparable from here
  on (Section~\ref{sec:data-quality}).
\end{itemize}

In short: the field has improved sharply. Completion on the corrected benchmark has roughly doubled, and the cost per task of a model with an equivalent task completion rate has dropped a hundredfold since 2024. The model results come first; the appendices document the updates to the
benchmark, the corrections, and engineering improvements that underlie
those numbers.

\section{Results}
\label{sec:results}

\subsection{Experimental Setup}
\label{sec:setup}
\textbf{Agent harness.} The original paper ran each model as a ReAct
\citep{yao2022react} loop that parsed a tool call out of free-form text. We instead
use the native tool-calling (structured output) interface that every current model
provider exposes, so the model emits a typed call against the tool schema and the
harness never has to recover a malformed action from prose. This single change removes the format-adherence failures that dominated the 2024 results (Section~\ref{sec:error-revisit}), and we adopt it for every model so
that the comparison is like-for-like. The loop is otherwise unchanged: the agent is given the task, all 26 tools, and up to 20 steps to reach a final state, with temperature set to zero where the model permits it. 

\textbf{Models.} We evaluate 24 models released between March 2023 and June 2026:
the GPT line from GPT-3.5-turbo through GPT-5.5, four Claude models (Fable 5,
Opus 4.8, Sonnet 4.6, and Haiku 4.5), two Gemini models (3.1-pro and 3.5-flash),
and six open-weight models (Qwen, DeepSeek, Kimi, GLM, Mistral Small 4, and
Mistral Medium 3.5).

\textbf{Cost estimation.} We estimate the
cost of one full benchmark run from the logged prompts and completions. Input and
output token counts are approximated from string lengths at four characters per
token, and the fixed per-call overhead of the system prompt and the full tool
schema (about 7{,}000 tokens, re-sent on every call) is added back. We price these
at each provider's published standard per-token rate, without caching. The
resulting figure is therefore an upper bound: a provider that caches the repeated
system prompt and schema would bill materially less.

\subsection{Findings}
For the 24 models, Table~\ref{tab:model-results} reports: successful task
completion, the rate of harmful side effects (a wrong action taken, such as an
email sent to the wrong person, where lower is better), and the estimated cost per
task.

\begin{table}[ht]
\setlength{\tabcolsep}{8pt}
\renewcommand*{\arraystretch}{1.25}
\centering
\begin{tabular}{l c c r}
\toprule
\textbf{Model} & \textbf{\makecell{Successful task\\completion ($\uparrow$)}}
& \textbf{Side effects ($\downarrow$)} & \textbf{Cost per task ($\downarrow$)} \\
\hline
Claude Fable 5     & 97.7\% & 1.9\%  & \$0.355 \\
Claude Opus 4.8    & 96.2\% & 2.5\%  & \$0.182 \\
Gemini-3.1-pro     & 95.5\% & 2.9\%  & \$0.076 \\
GPT-5.5            & 94.9\% & 3.9\%  & \$0.206 \\
Gemini-3.5-flash   & 92.0\% & 3.0\%  & \$0.067 \\
Claude Sonnet 4.6  & 88.3\% & 8.8\%  & \$0.105 \\
Kimi-K2.6          & 88.0\% & 6.8\%  & \$0.022 \\
GPT-5              & 84.3\% & 13.0\% & \$0.050 \\
DeepSeek-V4-pro    & 84.2\% & 12.8\% & \$0.017 \\
GPT-5.4            & 78.6\% & 16.8\% & \$0.087 \\
o3                 & 77.4\% & 17.5\% & \$0.072 \\
GLM-4.6            & 77.4\% & 17.0\% & \$0.017 \\
Claude Haiku 4.5   & 74.8\% & 16.7\% & \$0.034 \\
GPT-4.1            & 73.5\% & 19.4\% & \$0.065 \\
Qwen3.5            & 70.7\% & 21.4\% & \$0.003 \\
GPT-5.2            & 70.4\% & 18.8\% & \$0.055 \\
GPT-4o             & 69.0\% & 15.1\% & \$0.068 \\
Mistral Medium 3.5 & 61.4\% & 30.4\% & \$0.068 \\
GPT-4-turbo        & 61.3\% & 22.3\% & \$0.307 \\
GPT-5.1            & 60.0\% & 18.1\% & \$0.036 \\
GPT-5.4-mini       & 58.8\% & 30.3\% & \$0.027 \\
GPT-5.4-nano       & 48.6\% & 28.6\% & \$0.007 \\
Mistral Small 4    & 38.3\% & 44.3\% & \$0.006 \\
GPT-3.5-turbo      & 31.3\% & 38.7\% & \$0.016 \\
\bottomrule
\end{tabular}
\caption{\textbf{Task completion, side effects, and cost per task.} Completion and
side effects are over all 690 tasks; cost per task is the total benchmark spend
divided by 690. Rows are ordered by task
completion.
$\uparrow$ means higher is better and $\downarrow$ means lower is better.}
\label{tab:model-results}
\end{table}

\textbf{Completion has roughly doubled.} The best agent in 2024 completed 43\% of
tasks; the best in 2026, Claude Fable 5, completes 97.7\%. Nine models from five providers now clear
80\%. The strongest model still fails one task in forty. The failures it has left are mostly the harder reasoning and
multi-step retrieval cases.

\begin{figure}[t]
\centering
\includegraphics[width=0.85\linewidth]{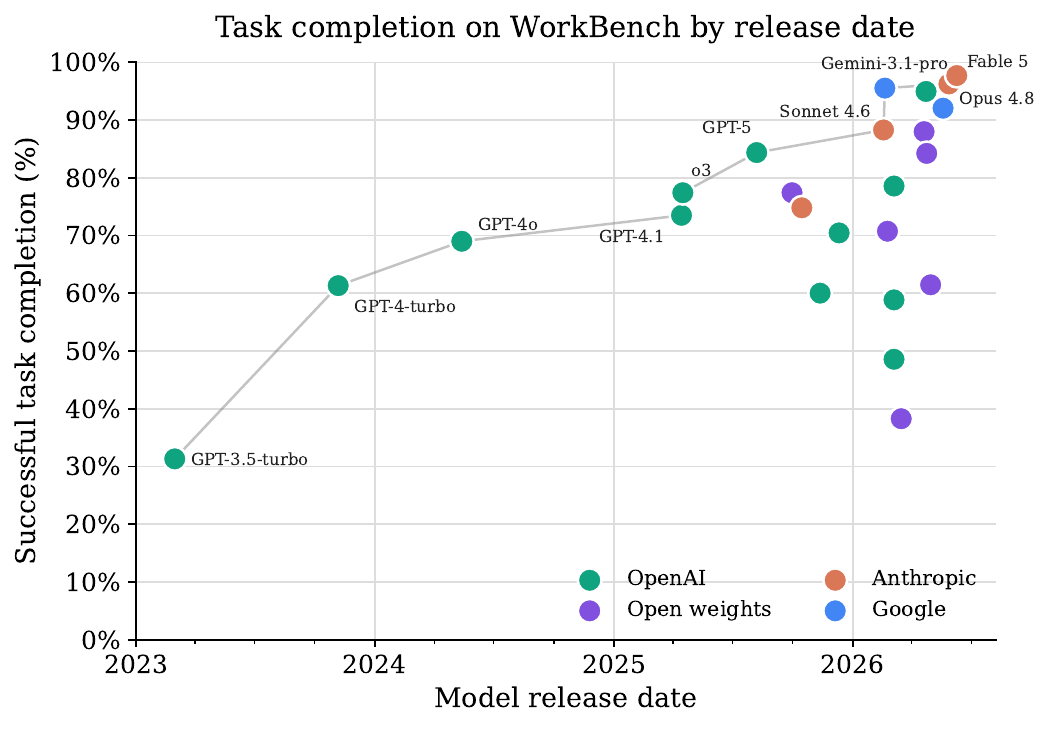}
\caption{\textbf{Task completion on WorkBench by release date.} Successful task
completion for every evaluated model against its public release date, coloured by
vendor. The grey line is the frontier of best completion over time: it only moves
upward, so models that fall below the running best are not joined to it.}
\label{fig:completion-vs-release}
\end{figure}

\textbf{Progress has been smooth since 2024.}
Figure~\ref{fig:completion-vs-release} plots task completion against each model's
public release date. The grey line traces the frontier, the best
completion reached by any model up to that date, and it climbs steadily from
GPT-3.5-turbo's 31.3\% in early 2023 to Fable 5's 97.7\% in mid-2026. Many later
releases sit well below that frontier: the smaller GPT-5.4-mini and -nano tiers, and
even some flagships such as GPT-5.1 and Mistral Medium 3.5, land below older models.

\begin{figure}[t]
  \centering
  \includegraphics[width=0.78\linewidth]{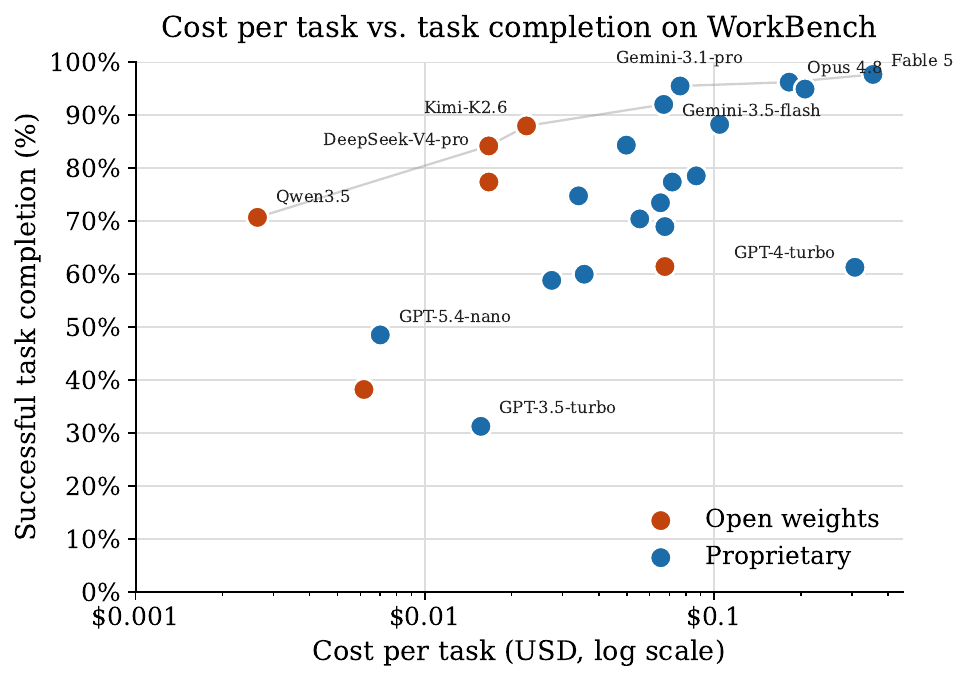}
  \caption{\textbf{Cost per task versus task completion on WorkBench.} Cost per task
  is the total spend to run the benchmark once divided by 690 tasks; the horizontal
  axis is logarithmic. Points are coloured by whether the model has open weights or is
  proprietary. The grey line is the efficient frontier (the most capable model at each
  price); only frontier models and a few notable outliers are labelled.}
  \label{fig:cost-vs-accuracy}
  \end{figure}

\textbf{Cost spans two orders of magnitude.}
Figure~\ref{fig:cost-vs-accuracy} plots cost per task against task completion on a
log cost axis. The grey line is the efficient frontier: its cheap end is entirely open weights,
and every one of those models comes from a Chinese lab: Qwen3.5,
DeepSeek-V4-pro, and Kimi-K2.6. The expensive, high-completion end is the mirror
image, entirely Western and proprietary: the two Gemini models, Opus 4.8, and Fable 5. What
falls off the frontier in between is mostly the Western budget tier. Claude Haiku 4.5
and the GPT-5.4 mini and nano models are neither the cheapest nor the most capable,
and each is dominated by a cheaper, more capable open-weight model: GPT-5.4-nano
(\$0.007, 49\%) loses outright to Qwen3.5 (\$0.003, 71\%), and Haiku 4.5
(\$0.034, 75\%) is beaten by DeepSeek, Kimi, and GLM at once. The
cheapest capable agent today is a Chinese open-weight model and the most capable is a
Western proprietary one. Western budget options are squeezed from both sides. 
The cost reduction over time is stark: Qwen3.5 exceeds the performance of GPT-4, the 2024 state-of-the-art,
at approximately one-hundredth the cost.

\section{Revisiting the Original Error Categories}
\label{sec:error-revisit}
With current models, we revisit the six largest sources of error from the original paper. The two largest sources for the frontier model of early 2024 (GPT-4) have
been largely eliminated by today's best models.

\begin{itemize}[leftmargin=*]
\itemsep0.3em
  \item \textbf{Failing to follow ReAct.} Eliminated. This is due to more
  tool-use-specific training and to constrained decoding (also
  known as structured outputs) to guarantee schema adherence.
  \item \textbf{Sending information to the wrong email address.} Almost eliminated.
  Models no longer treat the \texttt{name@example.com} in the docstring as a pattern
  to imitate, and instead resolve the correct address using the provided tool. We
  did, however, observe a single instance of GPT-5.5 sending an email to an
  \texttt{@example} address.
  \item \textbf{Failing to identify an available calendar slot.} Eliminated.
  Frontier LLMs consistently use the calendar search tool correctly and book the
  correct slot.
  \item \textbf{Misinterpreting retrieved data.} Reduced, though models still make
  basic errors. For example, given the task \textit{``please check the percent
  growth of engaged users since Friday. If it grew by more than average session
  duration\dots''}, Opus 4.8 compared the percentage growth of engaged users against
  the most recent raw value of average session duration.
  \item \textbf{Updating the wrong event.} Eliminated. Frontier LLMs consistently
  use the correct search tools to identify which event to update.
  \item \textbf{Using search incorrectly.} Reduced. Many search tools in WorkBench
  cap results at five per query, so the agent is expected to make multiple calls when
  more than five results are needed to complete a task. This still occasionally trips
  up frontier LLMs.
\end{itemize}

\section{Discussion}
\label{sec:discussion}
Two years ago, nothing cleared WorkBench. The best agent completed just 43\% of tasks. Today, the best agents largely pass it. It still discriminates, though: completion runs from 31\% to 98\%, cost ranges over two
orders of magnitude, and side effects range from 2\% to 44\%.

Cost is the axis with the most practical bite. A full run ranges from well under a
cent per task to over thirty cents, and the cheapest capable agents are now
open-weight models from Chinese labs rather than the proprietary frontier. Buying the
last few points of completion costs roughly an order of magnitude more per task than
settling just below the top of the table.

Further work could build upon WorkBench in several ways. For example, the sandbox is simpler than a real
workplace: a real inbox has thousands of historic messages and spam, and a real
calendar has years of events, so absolute completion here is an optimistic estimate
of deployed reliability. The cost figures are no-caching upper bounds derived from
token-length estimates rather than billed spend, and providers that cache the
repeated system prompt and tool schema would pay less, so cost should be read as an
ordering rather than an invoice. Outcome-centric evaluation, the feature that makes
WorkBench cheap and reproducible to score, can't grade pure-retrieval question
answering that leaves the sandbox unchanged. Finally, the
release-date analysis mixes vendors with different training objectives and data, so
the frontier in Figure~\ref{fig:completion-vs-release} is descriptive, not a claim
about any single lab's rate of progress.

WorkBench has been public on GitHub since
2024, and we keep no private hold-out set, so every task and its answer key is fair
game for a web crawl. Any model trained after the 2024 release may well have seen the
benchmark during pre-training. We have no clean way to tell
genuine capability gains from memorisation. A held-out split that's never published is an obvious improvement for the next iteration, and until there is one, the over-time results should be read as an upper bound on real progress.

\label{sec:conclusion}
WorkBench in 2026 is largely solved by frontier models: task completion on the corrected benchmark has
roughly doubled since 2024, the strongest agents now fail almost entirely in harmless
ways, and capability and safety improve together rather than trade off. Costs span two orders of magnitude, however, and cheaper models lag behind the frontier. We release the updated benchmark and harness,
and the per-model cost estimates so that the next generation of models can be placed
on the same axes.

\appendix
\section{Updates Since 2024}
\label{sec:data-quality}
The numbers above are scored on a benchmark that has changed since the 2024 release.
We have made two kinds of change: corrections that fix cases where the benchmark was
unfair to the agent, and engineering improvements to
the tools and task design. We document both here and quantify how much the
corrections move scores.

\subsection{Benchmark Corrections}
These corrections remove cases where the benchmark was unfair to the agent
or scored the wrong thing. A few representative examples:

\begin{itemize}[leftmargin=*]
\itemsep0em
  \item \textbf{Off-by-one in ``last $N$ days'' ground truth.} The date cutoff was
  computed as \texttt{today $- N$} rather than \texttt{today $- N{+}1$}, which
  shifted the answer key for a handful of email and calendar tasks. Fixed and
  regenerated.
  \item \textbf{Charts may end on the current day (56 tasks).} Analytics
  ``plot \dots\ since {date}'' tasks have a gold answer that ends the chart on the
  last day with data (2023-11-29). A model that instead plots up to the sandbox's
  current day (2023-11-30) draws the same chart over the same data, but was marked
  wrong because the saved plot's filename encodes its end date. Scoring now accepts
  either end date for \texttt{create\_plot}; nothing else is affected.
  \item \textbf{Prompt and answer-key mismatches.} Several tasks asked one thing
  but were graded against another. A placeholder always displayed ``more than''
  while the answer was sometimes computed for ``less than'', and one email task
  had a graded subject that was unreachable from the wording. The agent was being
  marked wrong for following the instructions correctly. Prompts now match the
  branch that is graded.
  \item \textbf{A silent-zero aggregation bug.} A ``fewest overdue tasks'' task
  used an \texttt{idxmin} that dropped people with zero overdue tasks, so the
  correct answer could never be a zero-count person. Reworded and recomputed.
  \item \textbf{Push-back tasks now respect working hours (1 task).} A ``push back
  my first meeting by 2 hours'' task moved a meeting to end at 18:30, breaking the
  prompt's own ``no meeting ends after 6pm'' rule, so a model that correctly
  refused was scored wrong. The delay is now clamped to keep the meeting ending by
  6pm, changing only that one task.
  \item \textbf{Email bodies show real newlines (10 tasks).} Send-email prompts
  displayed the body with escaped \texttt{\textbackslash n} while the answer key
  parsed to real newlines, so a faithful copy stored literal backslash-n and
  failed all 10 send-email tasks. In retrospect, this was ambiguous, since the agent does not know how the newlines will be interpreted downstream.
  The displayed body now contains real line
  breaks, leaving a single answer (the gold column is unchanged).
  \item \textbf{Clearer tool descriptions.} Tool docstrings now enumerate the
  allowed values for enum parameters and document result limits, so the agent is
  told the rules of the sandbox up front rather than discovering them by trial.
  \item \textbf{Counting queries made solvable.} ``Assign to the person with the
  fewest or most tasks'' was unwinnable while \texttt{search\_tasks} capped results
  at five and no tool enumerates people. The cap on \texttt{search\_tasks} alone is
  raised to 200 so the agent can enumerate a board and aggregate; other searches
  keep the cap of five.
\end{itemize}

\subsubsection{Scope of impact}
\label{sec:scope}
The cleanest way to see how much the benchmark changed is to hold the model fixed and
re-score it on both versions. GPT-4, the 2024 frontier model, went from 49\% on the
old benchmark to 57\% on the corrected one, resampled the same way on both. That
eight-point jump is the benchmark getting fairer, not the model getting smarter, and
it is the number to keep in mind whenever a 2026 result sits next to a 2024 one.

The fixes land on a minority of tasks. The hard floor, tasks whose
ground truth or wording changed, is 56 tasks (8\%), and that count is
deterministic. Widen it to include the fixes that change whether a correct attempt is
allowed to pass, the CRM read-data corrections and the move to order-independent
grading, and it is roughly 90 to 95 tasks (about 14\%). Everything else scores the
same task the same way it always did. Either way the takeaway is the same: a model's
2026 score isn't directly comparable to its 2024 score, so the comparison we trust is
between models run under the 2026 benchmark.

\section{Specific Failure Cases}
\label{sec:failures}
A collection of agent failures from the 2026 runs, each showing a task and how a
model went wrong.

\subsection{Over-zealous: acts when the condition is false}
\textbf{Task.} \textit{``I think dmitri might have some overdue tasks. Can you check
and if so, book a half hour meeting with them called `Catch up on overdue tasks' at
the earliest time I'm free tomorrow.''}\newline
\textbf{Failure.} Strictly, Dmitri has no \emph{past-due} tasks: every task with a
due date before today (2023-11-30) is already marked ``Completed''. Two tasks are due
\emph{today} and still incomplete, so they are about to become overdue but are not
overdue yet. The conditional (``if so'') should evaluate to false and the correct
outcome is to take no action, but the model treats ``about to be overdue'' as overdue
and books the meeting anyway.

\subsection{Reasoning failure: comparing a percentage to a raw value}
\textbf{Task.} \textit{``please check the percent growth of engaged users since
Friday. If it grew by more than average session duration make a front-end backlog
task called `Improve average session duration' for kofi that's due next Friday and
schedule a 30 minute meeting called `Discuss engaged users' for us at the earliest
slot i'm free tomorrow.''}\newline
\textbf{Failure.} The task compares two percent-growth figures and acts only if one
exceeds the other. The model instead compares a percentage against a raw value and skips the action the gold answer requires. It concludes:
\textit{``The engaged users grew by 0\%, which is not more than the average session
duration.''}

\subsection{Trusting a truncated search result}
\textbf{Failure.} Search tools return \emph{at most five} results, and the docstring
says so. When five calendar events come back, the model assumes it has seen them all
and stops, rather than re-querying with tighter parameters to narrow the set. It then
answers from an incomplete view of the data.

\bibliographystyle{retro_style}
\bibliography{references}

\end{document}